\documentclass[conference]{IEEEtran}
\IEEEoverridecommandlockouts

\usepackage{cite}
\usepackage{amsmath,amssymb,amsfonts}
\usepackage[T1]{fontenc}
\usepackage{algorithmic}
\usepackage{graphicx}
\usepackage{textcomp}
\usepackage{xcolor}
\usepackage{subcaption}
\usepackage{makecell}
\usepackage{array}
\usepackage{booktabs}
\usepackage[letterpaper]{geometry}
\usepackage{float}
\usepackage{svg}
\usepackage{eso-pic}

\def\BibTeX{{\rm B\kern-.05em{\sc i\kern-.025em b}\kern-.08em
		T\kern-.1667em\lower.7ex\hbox{E}\kern-.125emX}}

\newgeometry{top=25.4mm, bottom=19.1mm, left=19.1mm, right=19.1mm}
\begin{document}
	
\title{Scenario Generation in Roundabouts with Adjustable Interaction Intensity}

\author{
    Li Li\textsuperscript{1}, 
    Till Temmen\textsuperscript{1}, 
    Tobias Brinkmann\textsuperscript{1}, 
    Björn Krautwig\textsuperscript{1},\\
    Markus Eisenbarth\textsuperscript{1}
    and Jakob Andert\textsuperscript{1}
}

\maketitle

\footnotetext[1]{\raggedright Chair of Mechatronics in Mobile Propulsion, RWTH Aachen University, Germany, \{li\_li, brinkmann, temmen, krautwig, eisenbarth, andert\}@mmp.rwth-aachen.de
}

\AddToShipoutPicture*{%
    \AtTextLowerLeft{%
        \raisebox{-3\baselineskip}[0pt][0pt]{%
            \makebox[\textwidth][c]{\footnotesize
            © 2026 IEEE. This is the author's version of a work that has been accepted for publication in IEEE ITSC 2026.}%
        }%
    }%
}

\begin{abstract}
Roundabouts, characterized by frequent merging and yielding interactions, remain a safety-critical corner case for the development and testing of intelligent driving functions. However, extracting sufficient near-critical scenarios from naturalistic data is inefficient. Most existing scenario generation methods provide limited controllability over interaction intensity and criticality, making systematic safety testing and detailed analysis difficult. This paper presents an interaction-aware roundabout scenario generator with continuously adjustable interaction intensity. Geometric routes and temporal progress profiles are first decoupled and mapped to latent codes using pretrained autoencoders. Conditional latent generation is then performed with Wasserstein Generative Adversarial Networks (WGAN) to generate scenarios. Yielding is modeled as a controllable timing intervention via a compact yield code during the approach-to-entry segment, where interaction intensity is modulated by scaling the code with a factor $\lambda$. Results demonstrate enhanced timing-latent fidelity and plausible interaction responses compared to a baseline model. Under criticality-calibrated scaling, increasing $\lambda$ expands the safety margin, providing a scalable and controlled testing mechanism.
\end{abstract}

\section{Introduction}

Intelligent driving functions enhance driving convenience and comfort while becoming prevalent in production vehicles~\cite{chu2026imperfect_adas}. Meanwhile, rigorous validation is increasingly important to ensure reliable operation across diverse traffic situations, particularly in rare and safety-critical scenarios. Real-world road testing is a common validation approach in which vehicles are evaluated on public roads. However, due to high hardware costs and substantial time and staffing requirements, on-road testing alone cannot cover sufficient scenarios. Safety-critical interactions occur infrequently in naturalistic driving data and are expensive to reproduce at scale on public roads~\cite{kalra2016driving_to_safety}. Scenario-based testing in virtual environments has therefore become a cornerstone for safety assurance of intelligent driving functions, as it enables scalable execution, systematic variation, and controlled replay of critical scenarios without safety risk. Beyond safety assessment, realistic and controllable scenarios are also important for learning-based function development, where simulation and synthetic data can complement limited real-world corner cases~\cite{chen2024data_driven_traffic_sim_review}. 

While knowledge-based scenario generation is interpretable and easy to configure, its scalability is limited by low generation efficiency and insufficient coverage of corner cases. Consequently, data-driven methods are gaining traction due to their higher efficiency and improved realism. Such methods utilize real-world data together with generative techniques for scenario creation. A key challenge is that many data-driven scenario generators primarily optimize distributional fidelity, whereas control over interaction outcomes remains limited or entangled with confounding factors~\cite{chen2024data_driven_traffic_sim_review,ding2023survey_scengen}. However, function development and validation require more targeted testing capabilities, particularly to characterize the failure boundaries of functions such as Automated Emergency Braking (AEB) and Collision Avoidance (CA). This is especially relevant in terms of false activation and insufficient intervention under near-critical traffic situations~\cite{nhtsa2024_ncap_adas_roadmap,fu2020_aeb_sdt}.

Roundabouts are widely deployed at-grade intersections and constitute a challenging operational design domain due to curved geometry, sustained steering and frequent acceleration/deceleration demands. Circulating vehicles have right of way and entering vehicles must yield, which poses risk during the approach-to-entry segment. Empirical evidence and recent reviews highlight that yield-related interaction failures are a prominent safety concern at roundabouts, motivating targeted and scalable testing of yielding and gap-acceptance behaviors in virtual environment \cite{li2023roundabout_conflict_review,Polders2013Crash}.

This paper presents an interaction-aware scenario generation framework for roundabouts with continuously adjustable interaction intensity. The framework first decouples geometric routes and along-route progress profiles from naturalistic trajectories, which are then mapped into latent spaces via pretrained autoencoders. Wasserstein Generative Adversarial Networks with Gradient Penalty (WGAN-GP) is employed to perform conditional latent synthesis for scenario generation \cite{arjovsky2017wgan,gulrajani2017wgangp}. Interaction is represented by a compact yield code whose intensity is scaled by a factor $\lambda$, enabling controlled changes in the entering vehicle's temporal behavior. The resulting interaction effect is quantified at the scenario level, thereby facilitating criticality-calibrated scaling for systematic testing.

In summary, the main contributions of this paper are as follows:
\begin{itemize}
    \item A three-stage roundabout scenario generator that decouples route geometry and temporal profiles via pretrained autoencoders, and synthesizes latent space using conditional WGAN-GP.
    \item A yielding interaction representation with a continuous intensity factor $\lambda$, enabling adjustable interaction of the entering vehicle.
    \item A criticality-calibrated testing protocol that identifies a near-critical reference via trajectory alignment, facilitating systematic safety evaluations by scaling interaction intensity under controlled conditions.
\end{itemize}

\section{Related Work}

\subsection{Traffic Scenario Generation}
Scenario-based validation has motivated extensive research on efficient test-case generation for intelligent driving functions. A representative line of work formulates testing scenario library generation (TSLG), where scenarios are selected and organized to balance exposure frequency and maneuver challenge. This enables accelerated evaluation compared with brute-force on-road testing \cite{feng2022tslg_adaptive}. Complementarily, naturalistic-and-adversarial environment models aim to reduce the required test miles by learning how background vehicles should act to expose rare safety-critical events \cite{feng2021idit,liu2024curse}. These results highlight sampling efficiency as a central objective in large-scale safety validation.

Data-driven generative approaches synthesize realistic scenarios directly from real-world data. Scenario-parameter generation and representativeness metrics have been proposed to quantify coverage and realism in scenario-based assessment \cite{degelder2022sr}. 
CVAE-T was proposed for multi-agent scenario generation in roundabouts, and its latent space was analyzed for partial disentanglement and interpretable effects on entry timing and velocity profiles~\cite{li2025multi}. For safety testing under limited critical samples, WGAN-GP has been used to generate additional risky trajectory fragments \cite{li2023aap_wgangp}. Transformer time-series GANs further improve long-horizon dependency modeling for pre-crash trajectory generation \cite{liu2025eaai_ttsgan}.
Recent work has also explored diffusion-based driving scenario generation with controllable sampling to create realistic and safety-critical traffic scenes \cite{xu2025diffscene}, and the use of evolving interactive background agents to test automated vehicles under human-like multi-agent interactions \cite{ma2024evolving_trc}. Despite these advances, many existing generators entangle route geometry, timing, and interaction within a joint learning paradigm. This coupling complicates the isolation of safety impacts stemming from specific interaction mechanisms under controlled conditions. From a model-selection perspective, comparative surveys indicate that Variational Autoencoder (VAE) objectives can bias generation toward averaged samples, while diffusion models typically require an iterative reverse-time sampling procedure. Therefore, a latent-space WGAN-GP is adopted here to enable one-pass conditional sampling that fits into a criticality-calibrated testing protocol with a limited simulation budget \cite{bondtaylor2022dgm,cao2024diffusion}.

\subsection{Surrogate Safety Measures and Safety Testing}

For safety evaluation, surrogate safety measures (SSMs) quantify near-miss severity without requiring crash observations. Time-to-collision (TTC) and post-encroachment time (PET) are among the most widely used measures \cite{hayward1972near_miss,peesapati2018pet,astarita2019surrogate}. For roundabout approach-to-entry segment, gap-based time margins between entering and circulating streams are commonly used to characterize yielding demand and safety margins \cite{memeh2024roundabout_gap_crash,li2023roundabout_conflict_review}. However, existing interaction descriptors and SSMs are primarily used for assessment or rule checking, and are less explored as continuous control variables inside data-driven generators to enable controllable interaction variation.

Safety testing aims to discover near-miss or failure cases efficiently under a limited simulation budget. A major direction is accelerated evaluation via rare-event simulation techniques such as importance sampling to increase the frequency of safety-critical events while preserving statistically meaningful estimates \cite{zhao2018carfollow}. Adaptive sampling strategies further concentrate tests near performance boundaries or informative regions of the parameter space, for example via adaptive design of experiments \cite{sun2022doe}.

Another complementary direction directly searches for hazardous or boundary cases via optimization and iterative refinement. Optimization-search-based scenario enhancement can amplify hazardousness while respecting feasibility constraints \cite{zhu2022hazard}. These methods emphasize that a useful testing pipeline should not only generate realistic scenarios, but also provide mechanisms to vary the criticality by adjusting of interaction behavior, which aligns with the goal of this work.

\section{Methodology}

\subsection{Data Processing and Preparation}
\label{sec:method_data}

\begin{figure}[htbp]
    \centering
    \includegraphics[width=0.48\textwidth]{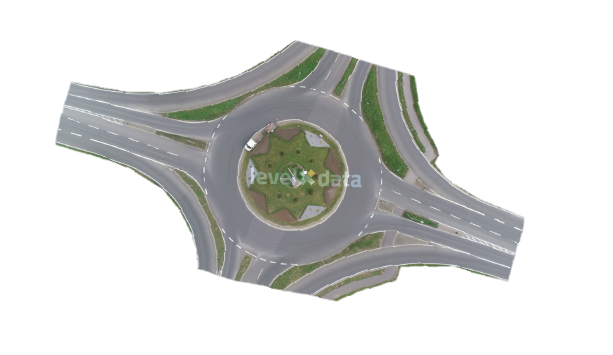}
    \caption{Neuweiler Roundabout in rounD Dataset \cite{krajewski2020round}.}
    \label{fig:roundabout}
\end{figure}

The Neuweiler roundabout from the rounD dataset is selected because of its well-structured data and representative four-sector layout. Trajectories are downsampled by a factor $r=3$ to accelerate training, resulting in a uniform step size $dt=0.12\,\mathrm{s}$. The start and end positions of each vehicle are categorized into one of four sectors, defining the entry-exit condition $c =(c_{\text{in}},c_{\text{out}})\in\{1,2,3,4\}^2$.

As vehicle routes and speeds vary across samples, the valid trajectory frame length $L$ is sample-dependent. Let $\mathbf{x}(t)\in\mathbb{R}^2$ denote the vehicle position at discrete index $t$ in the global $x$--$y$ coordinate system. Over the valid segment $t=0,1,\ldots,L-1$, the cumulative route length is defined by
\begin{equation}
    \begin{aligned}
        s(0) &= 0, \\
        s(t) &= \sum_{k=1}^{t} \left\lVert \mathbf{x}(k) - \mathbf{x}(k-1) \right\rVert_{2}, \quad t = 1, \dots, L-1.
    \end{aligned}
    \label{eq:arc_def}
\end{equation}

The normalized progress variable is defined as
\begin{equation}
    u(t)=\frac{s(t)}{s(L-1)}, \qquad t=0,1,\ldots,L-1,
    \label{eq:progress_def}
\end{equation}
where $s(L-1)$ is the total route length $l$. By construction, $u(0)=0$ and $u(L-1)=1$ when $l>0$. The geometric route is resampled on a fixed progress grid of $M=128$ points, $u_m=m/(M-1)$ for $m=0,1,\ldots,M-1$. Linear interpolation is employed to construct the geometric sequence $\boldsymbol{\gamma} = \{\mathbf{x}_m\}_{m=0}^{M-1}$, where each $\mathbf{x}_m$ represents the spatial coordinates at progress $u_m$. The progress sequence is padded to a fixed frame length $T=234$ by repeating the last valid value. The normalized valid frame length is caculated by $\ell=(L-1)/(T-1)\in[0,1]$. The normalized total route length is denoted by $\bar l\in[0,1]$, obtained from $l$ via min--max normalization over the training set. Given the resampled geometric route $\boldsymbol{\gamma}=\{\boldsymbol{\gamma}_m\}_{m=0}^{M-1}$ on the fixed progress grid $\{u_m\}$ and a progress value $u(t)\in[0,1]$, the corresponding vehicle position can be recovered by linear interpolation along the route. 

This preprocessing decouples each trajectory into fixed-length geometric routes and timing sequences. The resulting consistent dimensions facilitate stable training while allowing the model to capture spatial and temporal features separately. All processed data are summarized in Table~\ref{table_sequences} and split into disjoint training/validation/test sets with a 0.70/0.15/0.15 proportion.

\begin{table}[!t]
    \renewcommand{\arraystretch}{1.3}
    \caption{Processed Variables and Data Dimensions}
    \label{table_sequences}
    \centering
    \begin{tabular*}{0.95\columnwidth}{@{\extracolsep{\fill}} c l l @{}}
        \hline
        \textbf{Variable} & \textbf{Space} & \textbf{Description} \\
        \hline
        $\boldsymbol{\gamma}$ & $\mathbb{R}^{N\times M\times 2}$ & Geometric sequence \\
        $\mathbf{u}$ & $\mathbb{R}^{N\times T\times 1}$ & Progress sequence \\
        $\ell$   & $\mathbb{R}^{N}$           & Normalized valid frame length \\
        $\bar l$      & $\mathbb{R}^{N}$          &  Normalized total route length \\
        $\mathbf{c}$     & $\{1,2,3,4\}^{N\times 2}$         & Geometric conditions \\
        \hline
    \end{tabular*}
\end{table}

\subsection{Interaction Behavior in Roundabout}
\label{sec:method_yield}

The failure to yield, where an entering vehicle insufficiently accounts for circulating traffic, remains a key factor to roundabout accidents and serves as the primary focus of this study.~\cite{Polders2013Crash}. Yield demand is evaluated on the approach-to-entry segment using an arrival-time proximity (ATP) with respect to a fixed crossing point $p^\star$ at the entry. $p^\star$ is defined as the intersection of the entry-lane and circulating-lane centerlines. Let $\Delta s_e(t)>0$ denote the ego distance to $p^\star$ along the entry-lane centerline and $v_e(t)$ the corresponding speed. For each circulating candidate $k\in\mathcal K(t)$ (vehicles on the circulating lane within $40\,\mathrm{m}$ upstream of $p^\star$ along the circulating-lane centerline), let $\Delta s_k(t)$ and $v_k(t)$ be defined analogously. The estimated times-to-arrival are
\begin{equation}
\hat{t}_e(t)=\frac{\left[\Delta s_e(t)-r_{\mathrm{veh}}\right]_+}{\max(v_e(t),\delta_v)},\quad
\hat{t}_k(t)=\frac{\left[\Delta s_k(t)-r_{\mathrm{veh}}\right]_+}{\max(v_k(t),\delta_v)},
\end{equation}
where $\delta_v=10^{-3}\,\mathrm{m/s}$, $r_{\mathrm{veh}}=2\,\mathrm{m}$, and $[x]_+=\max(x,0)$. The signed arrival-time gap is $\Delta t_k(t)=\hat{t}_k(t)-\hat{t}_e(t)$. As non-negative criticality scoring, the ATP is defined:
\begin{equation}
\mathrm{ATP}(t)=
\begin{cases}
\min\limits_{k\in \mathcal{K}(t)} \left| \Delta t_k(t) \right|, & \Delta s_e(t)>0 \ \text{and}\ \mathcal{K}(t)\neq\emptyset,\\
\mathrm{ATP}_{\max}, & \text{otherwise},
\end{cases}
\end{equation}
where $\mathrm{ATP}_{\max}=6\,\mathrm{s}$. Yield demand is active at time $t$ if $\Delta s_e(t)>0$, $\mathcal{K}(t)\neq\emptyset$ and $\mathrm{ATP}(t)\le 6\,\mathrm{s}$. The minimum-ATP (minATP) is defined as $\min_{t}\mathrm{ATP}(t)$ over approach-to-entry steps. The time index $t^\star=\arg\min_t \mathrm{ATP}(t)$ is used to compute clearance at the minATP moment. Clearance is defined as the Euclidean distance between the two vehicles minus a $4\,\mathrm{m}$ offset (a coarse tolerance for vehicle size).

A fixed-length yield code $\mathbf{y}$ is constructed for each trajectory based on aligned per-step spatial and temporal data extracted from the dataset. Using the binary yield-demand signal and the associated ATP on valid steps, four scalars are computed:
\begin{itemize}
    \item $y_{\mathrm{pres}}$: a binary indicator that equals 1 if at least one valid step is flagged as yield demand, and 0 otherwise.
    \item $y_{\mathrm{frac}}$: the fraction of valid steps that are flagged as yield demand.
    \item $y_{\mathrm{minATP}}$: the normalized minATP value by a constant $\mathrm{ATP}_{\max}$ and clipped to $[0,1]$. It is set to 1 when no yield-demand step exists.
    \item $\tau_{\mathrm{peak}}$: the normalized time location of the most critical yield moment, defined as the index of the yield-demand step with the minATP divided by $(L-1)$. It is set to 0 when $y_{\mathrm{pres}}=0$.
\end{itemize}

A neutralized yield code $\mathbf{y_n}$ is formed by setting the yield dimensions to neutral values:
\begin{equation}
\begin{aligned}
    &y^{(i,0)}_{\mathrm{pres}}=0, && y^{(i,0)}_{\mathrm{frac}}=0, \\
    &y^{(i,0)}_{\mathrm{minATP}}=1, && \tau_{\mathrm{peak}}^{(i,0)} = 0.
\end{aligned}
\label{eq:yield_neutral}
\end{equation}

\subsection{Autoencoder and WGAN-GP Models}
\label{sec:method_wgan}

\begin{figure}[htbp]
    \centering
    \includegraphics[width=0.48\textwidth]{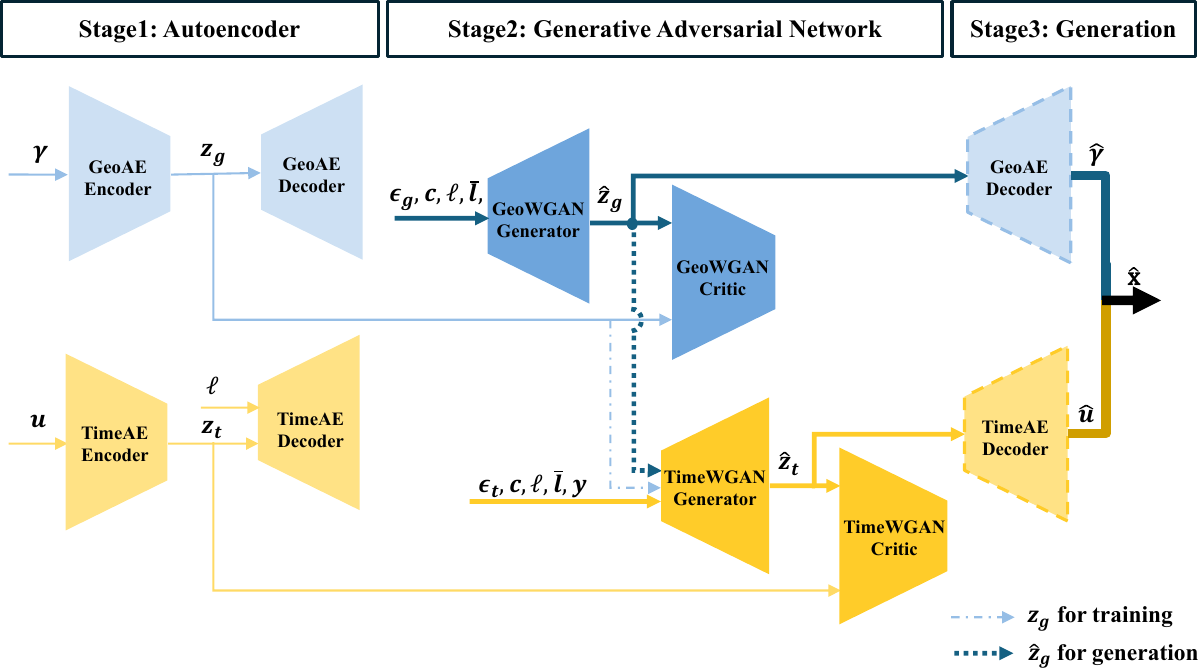}
    \caption{Overview of the Three-Stage Scenario Generation Pipeline.}
    \label{fig:framework}
\end{figure}

As shown in Fig.~\ref{fig:framework}, GeoAE and TimeAE are pretrained in stage 1 to reduce dimensionality and stabilize adversarial training, and then frozen for decoding in Stage~3. Given a geometric sequence $\boldsymbol{\gamma}$ and a progress sequence $\mathbf{u}$, the encoders produce latent codes
\begin{equation}
    \mathbf{z}_g = E_g(\boldsymbol{\gamma}), \qquad \mathbf{z}_t = E_t(\mathbf{u}),
\end{equation}
and the decoders reconstruct
\begin{equation}
    \hat{\boldsymbol{\gamma}} = D_g(\mathbf{z}_g), \qquad \hat{\mathbf{u}}=D_t(\mathbf{z}_t, \ell).
\end{equation}
In the implemented pipeline, the encoded latents are precomputed for all training samples and stored as arrays $\mathbf{z}_g$ and $\mathbf{z}_t$.

Latent generation is performed using two WGAN-GP models in stage 2. A GeoWGAN is trained in the GeoAE latent space to sample route latents. Its generator $G_g$ maps $(\boldsymbol{\boldsymbol{\epsilon}}_g,\mathbf{c},\ell, \bar l)$ to $\hat{\mathbf{z}}_g$ and critic $C_g$ is optimized under a standard WGAN-GP framework. The generated geometric route is obtained by decoding $\hat{\boldsymbol{\gamma}}=D_g(\hat{\mathbf{z}}_g)$. The TimeWGAN Generator additionally conditions on the geometry latent $\mathbf{z}_g$ and a yield code vector $\mathbf{y}$ in the training. The conditional latent generator is
\begin{equation}
    \hat{\mathbf{z}}_t = G_t(\boldsymbol{\boldsymbol{\epsilon}}_t, \mathbf{c}, \ell, \bar l, \mathbf{z}_g, \mathbf{y}), \qquad \boldsymbol{\boldsymbol{\epsilon}}_t \sim \mathcal{N}(0, I),
    \label{eq:zt_gen_method}
\end{equation}
and a critic $C_t(\cdot)$ scores real and generated timing latents under the same conditions. The WGAN-GP objective is
\begin{align}
    \mathcal{L}_{C_t} &=
    \mathbb{E}\!\left[C_t(\hat{\mathbf{z}}_t \mid \mathbf{c},\ell,\bar l,\mathbf{z}_g,\mathbf{y})\right]
    -
    \mathbb{E}\!\left[C_t(\mathbf{z}_t \mid \mathbf{c},\ell,\bar l,\mathbf{z}_g,\mathbf{y})\right]
    \nonumber\\
    &\quad + \lambda_{\mathrm{gp}}\,
    \mathbb{E}\!\left[\Big(\|\nabla_{\tilde{z}} C_t(\tilde{\mathbf{z}} \mid \mathbf{c},\ell,\bar l,\mathbf{z}_g,\mathbf{y})\|_2 - 1\Big)^2\right],
    \label{eq:wgangp_critic_method}
    \\
    \mathcal{L}_{G_t} &= -\,\mathbb{E}\!\left[C_t(\hat{\mathbf{z}}_t \mid \mathbf{c},\ell,\bar l,\mathbf{z}_g,\mathbf{y})\right],
    \label{eq:wgangp_gen_method}
\end{align}
where $\tilde{\mathbf{z}}=\alpha \mathbf{z}_t + (1-\alpha)\hat{\mathbf{z}}_t$ with $\alpha\sim U(0,1)$. Post-training, the temporal progress profile is obtained by decoding $\hat{\mathbf{u}}=D_t(\hat{\mathbf{z}}_t,\ell)$. The final spatiotemporal trajectory is then generated by interpolating the geometric route $\hat{\boldsymbol{\gamma}}$ according to the progress profile $\hat{\mathbf{u}}$.

\subsection{Interaction-aware Scenario Generation}
\label{sec:method_interaware}

Two sets of single-vehicle conditions from the dataset are sampled and used for generation. Initially, Vehicle~B is generated to represent the circulating vehicle. Subsequently, Vehicle~A, representing the entering vehicle, is generated under two distinct configurations. An Interaction-Off variant uses the neutralized yield code and an Interaction-On variant uses the corresponding yield code, while keeping identical geometric conditions and the same latent noise. 

A two-vehicle scenario is assembled by applying a global time shift $\Delta t$ to Vehicle~B relative to Vehicle~A (positive $\Delta t$ delays Vehicle~B). To obtain non-trivial scenarios, the initial $\Delta t$ is determined via a randomized grid search on the Interaction-Off baseline. This search identifies a $\Delta t$ that yields a scenario-level minATP within a predefined target band, where specific yielding behavior is required. The grid search is conducted over a uniform grid $\{\Delta t_j\}_{j=1}^{201}$ spanning $[-12\,\mathrm{s},\,12\,\mathrm{s}]$ in a randomly permuted order. The first $\Delta t$ yielding a minATP within the target band is selected. If no candidate satisfies, the one with the minATP most proximal to the interval is used. A fixed seed is employed for the random permutation to ensure reproducibility.

Once $\Delta t$ is selected, the aligned Interaction-Off scenario is used to extract the yield code for Vehicle~A. This code is then employed to generate the Interaction-On variant, while preserving the geometric conditions, latent noise, and time alignment. An interaction-intensity sweep is performed by scaling the yield code with a factor $\lambda\in[0,1]$. The interpolated code is defined as
\begin{equation}
\begin{aligned}
    &y_{\mathrm{pres}}^{(i,\lambda)} = \mathbb{I}[\lambda>0]\; y_{\mathrm{pres}}^{(i,1)}, && y_{\mathrm{frac}}^{(i,\lambda)} = \lambda y_{\mathrm{frac}}^{(i,1)}, \\
    &y_{\mathrm{minATP}}^{(i,\lambda)} = 1 - \lambda\bigl(1 - y_{\mathrm{minATP}}^{(i,1)}\bigr), \\
    &\tau_{\mathrm{peak}}^{(i,\lambda)} = \mathbb{I}[\lambda>0]\;\tau_{\mathrm{peak}}^{(i,1)}.
\end{aligned}
\label{eq:lambda_all}
\end{equation}
A lower $\lambda$ value reduces the effective interaction strength, thereby representing higher scenario criticality. 

In summary, scenario criticality is first calibrated by identifying the time shift $\Delta t$ using the Interaction-Off variant as reference. The yield code derived from this reference provides the necessary condition for interaction-aware scenario generation. The interaction-intensity parameter $\lambda$ scales the magnitude of the yielding intervention within the generated temporal progress profile of Vehicle~A.

\section{Experimental Setup}
\label{sec:exp}

\subsection{Model Training and Implementation Details}
\label{sec:exp_train}

Table~\ref{tab:ae_setup} summarizes the autoencoder architectures used in the implementation. Both autoencoders are trained with Mean Squared Error (MSE) and Adam (learning rate $10^{-3}$, batch size 64). Early stopping is applied on the validation loss with patience 20. The learning rate is reduced by a factor of 0.5 if the validation loss does not improve for 10 epochs, with a minimum learning rate of $10^{-6}$. GeoAE is trained up to 1000 epochs and TimeAE up to 2000 epochs. The TimeAE decoder predicts non-negative progress increments using a softplus transform and cumulative integration. The resulting sequence is then normalized by its endpoint and clipped to the $[0,1]$ range.

After training GeoAE and TimeAE, all trajectories are encoded once to obtain the latent codes $(\mathbf{z}_g,\mathbf{z}_t)$, which serve as real samples for adversarial training. GeoWGAN and TimeWGAN are then trained in the corresponding latent spaces, with the optimization settings summarized in Table~\ref{tab:gan_setup}. In each iteration, the critic is updated $n_{\mathrm{critic}}$ times per generator update, and a gradient penalty with weight $\lambda_{\mathrm{gp}}$ is applied. To mitigate yield-class imbalance, each TimeWGAN mini-batch is formed by stratified sampling with a fixed yield-present sample fraction $p_{\mathrm{yield}}=0.6$.

The implemented generator and critic in both WGANs are multilayer perceptrons with explicit conditioning embeddings. Conditions are mapped to trainable embeddings and concatenated. Then, all input data are mapped through small fully connected blocks and concatenated with the Gaussian noise. The WGAN Generator uses three fully connected blocks (256--512--512) with LeakyReLU activations and batch normalization. The WGAN Critic uses fully connected blocks (512--256--128) with LeakyReLU and outputs a scalar score. 

\begin{table}[t]
\caption{Autoencoder Architectures.}
\label{tab:ae_setup}
\centering
\footnotesize
\renewcommand{\arraystretch}{1.15}
\begin{tabular*}{0.95\columnwidth}{@{\extracolsep{\fill}}cccc}
\hline
Model & Encoder & Decoder & Latent dim. \\
\hline
GeoAE  & 256--128--64 & 64--128--256 & 64 \\
TimeAE & 234--128--64--16 & 16(+1)--64--128--234 & 16 \\
\hline
\end{tabular*}
\end{table}

\begin{table}[t]
    \caption{WGAN-GP Optimization Settings.}
    \label{tab:gan_setup}
    \centering
    \footnotesize
    \renewcommand{\arraystretch}{1.15}
    \begin{tabular*}{0.95\columnwidth}{@{\extracolsep{\fill}}l c}
        \hline
        Setting & Value \\
        \hline
        Epochs (GeoWGAN / TimeWGAN) & 1000 / 600 \\
        Batch size & 64 \\
        Critic updates per generator ($n_{\mathrm{critic}}$) & 5 \\
        Gradient penalty weight ($\lambda_{\mathrm{gp}}$) & 10 \\
        Adam (lr, $\beta_1,\beta_2$) & $5\times10^{-5}$, 0, 0.9 \\
        Global gradient norm clipping & 1.0 \\
        \hline
    \end{tabular*}
\end{table}

\subsection{Baselines and Evaluation Protocols}
To evaluate the contribution of interaction-aware modeling, two model configurations are compared:
\begin{itemize}
    \raggedright
    \item Baseline: a TimeWGAN trained on neutralized yield code $\mathbf{y_n}$.
    \item Interaction-aware: an interaction-aware TimeWGAN trained on authentic yield code $\mathbf{y}$.
\end{itemize}
Both variants share the same architecture and input interface. Using a shared GeoWGAN generator ensures that differences in interaction metrics primarily reflect timing changes rather than geometric variations. The baseline model is used for distributional fidelity comparisons. 1000 trajectories are generated from each variant by a fixed set of conditioning tuples from the test sets. Generated timing latents are compared to real timing latents using Maximum Mean Discrepancy (MMD) with a Radial Basis Function (RBF) kernel. The Fréchet distance is interpreted as a second-order moment matching metric rather than evidence of Gaussianity. 

Interaction-aware evaluation follows the procedure detailed in Sec.~\ref{sec:method_interaware} to compare the Interaction-Off reference with the Interaction-On variant. For each of the 100 synthesized scenarios, the minATP and the corresponding clearance at the minATP moment are computed for both cases. Results are reported as mean values within each predefined minATP band. Representative $x$--$y$ plots are utilized to visualize the aligned vehicle positions at key time steps.

For evaluating criticality-calibrated scaling, $\Delta t$ is obtained by a band-targeted grid search on the Interaction-Off such that the reference scenario matches a prescribed minATP target band $[0,2.0]$\,s. Ten reference safety-critical scenarios are uniformly generated without interaction behavior. Each calibrated reference is evaluated under $\lambda\in\{0,0.1,0.2,\ldots,1.0\}$, resulting in 110 points per scatter plot with color indicating $\lambda$. 

\section{Results}\label{result}

\subsection{Realism Evaluation}
\label{sec:results_main}

\begin{figure}[t]
    \centering
    \includegraphics[width=0.45\textwidth]{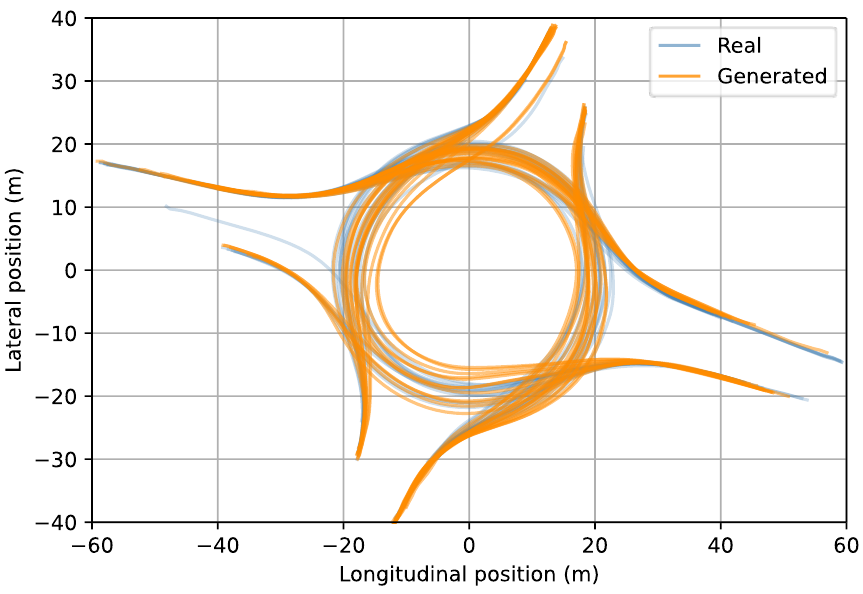}
    \caption{Overlay of Real and Generated routes.}
    \label{fig:geom_overlay}
\end{figure}

Fig.~\ref{fig:geom_overlay} illustrates an overlay of 50 generated trajectories against 50 real trajectories. The generated geometric routes closely adhere to the empirical distributions, suggesting that the geometry sampler effectively captures the dominant roundabout patterns. For a quantitative assessment, Table~\ref{tab:fidelity_baseline_vs_inter} presents distributional fidelity metrics across 1000 generated samples. The interaction-aware TimeWGAN significantly enhances timing-latent fidelity. The MMD in $\mathbf{z}_t$ reduced from 0.3997 to 0.1796, while Fr\'echet distance drops from 0.1894 to 0.0020. 

\begin{table}[t]
\caption{Comparison Between Baseline and Interaction-aware Model.}
\label{tab:fidelity_baseline_vs_inter}
\centering
\footnotesize
\setlength{\tabcolsep}{6pt}
\renewcommand{\arraystretch}{1.12}
\begin{tabular*}{0.95\columnwidth}{@{\extracolsep{\fill}} l c c @{}}
\hline
Metric & Baseline & Interaction-aware \\
\hline
MMD in $z_{\mathrm{t}}$ $\downarrow$ & 0.3997 & \textbf{0.1796} \\
Fr\'echet distance in $z_{\mathrm{t}}$ $\downarrow$ & 0.1894 & \textbf{0.0020} \\
\hline
\end{tabular*}
\end{table}

\subsection{Interaction-aware Scenario Generation}
\label{sec:results_interaware}

Following the procedure in Sec.~\ref{sec:method_interaware}, two-vehicle scenarios are generated to compare the Interaction-Off reference and the Interaction-On variant within specific minATP intervals. To ensure the inclusion of meaningful interactions, $\Delta t$ is selected such that the Interaction-Off scenario resides within a predefined minATP range of $[0,\,4]\,\mathrm{s}$.

\begin{figure}[t]
    \centering
    \begin{subfigure}{0.24\textwidth}
        \centering
        \includegraphics[width=\linewidth]{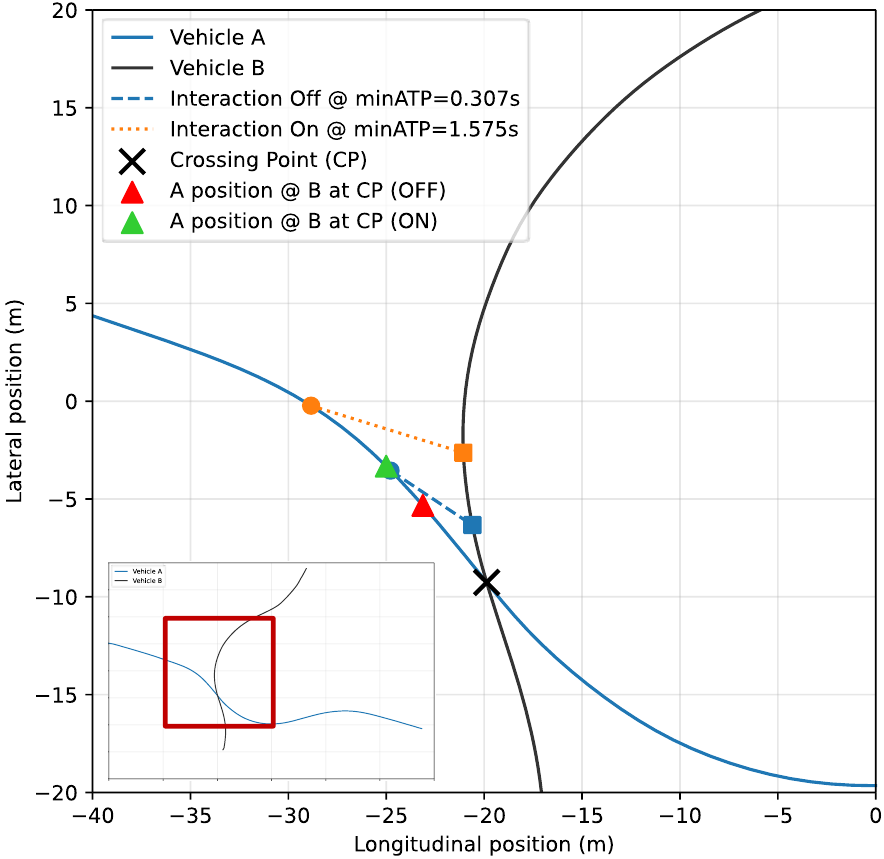}
    \end{subfigure}
    \hfill
    \begin{subfigure}{0.24\textwidth}
        \centering
        \includegraphics[width=\linewidth]{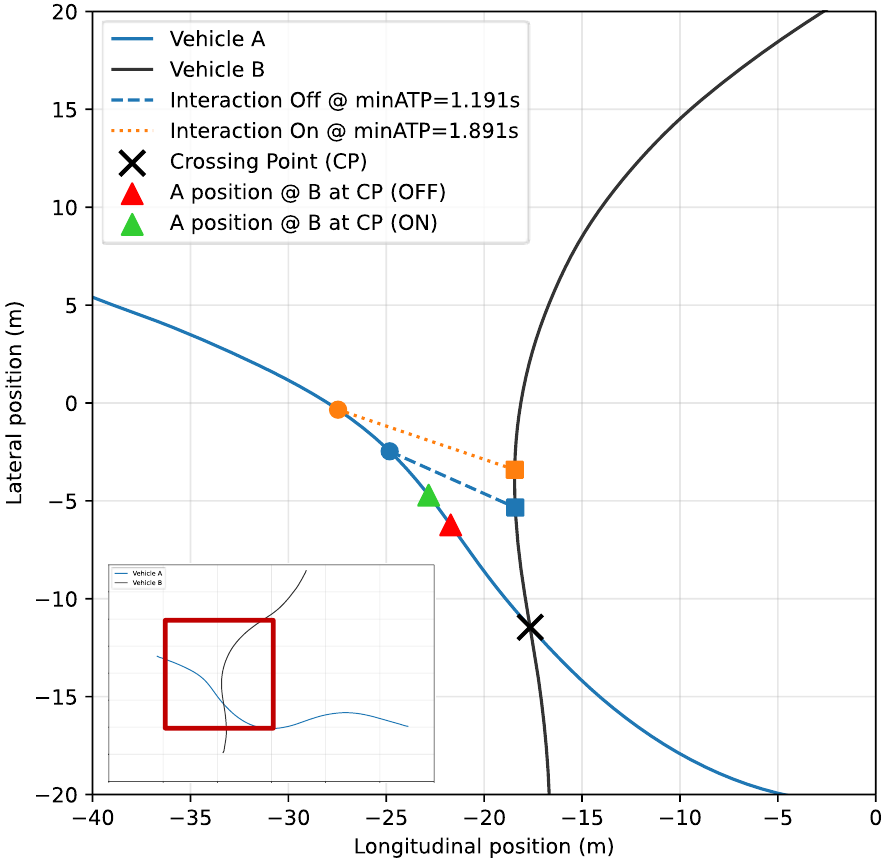}
    \end{subfigure}
    
    \vspace{2mm}

    \begin{subfigure}{0.24\textwidth}
        \centering
        \includegraphics[width=\linewidth]{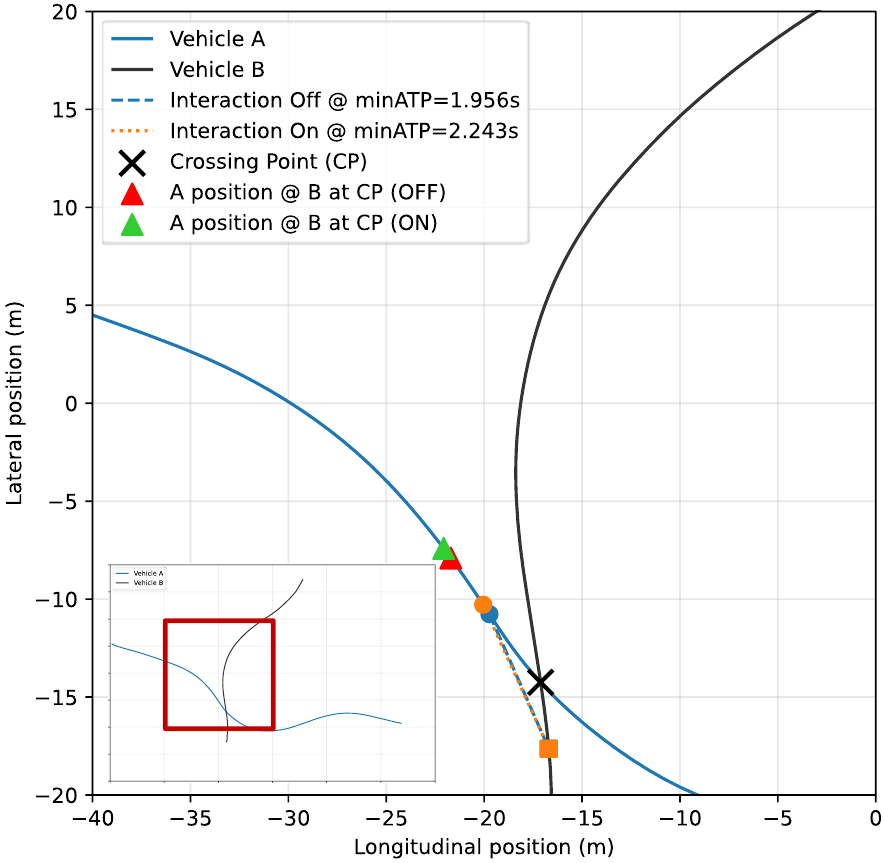}
    \end{subfigure}
    \hfill
    \begin{subfigure}{0.24\textwidth}
        \centering
        \includegraphics[width=\linewidth]{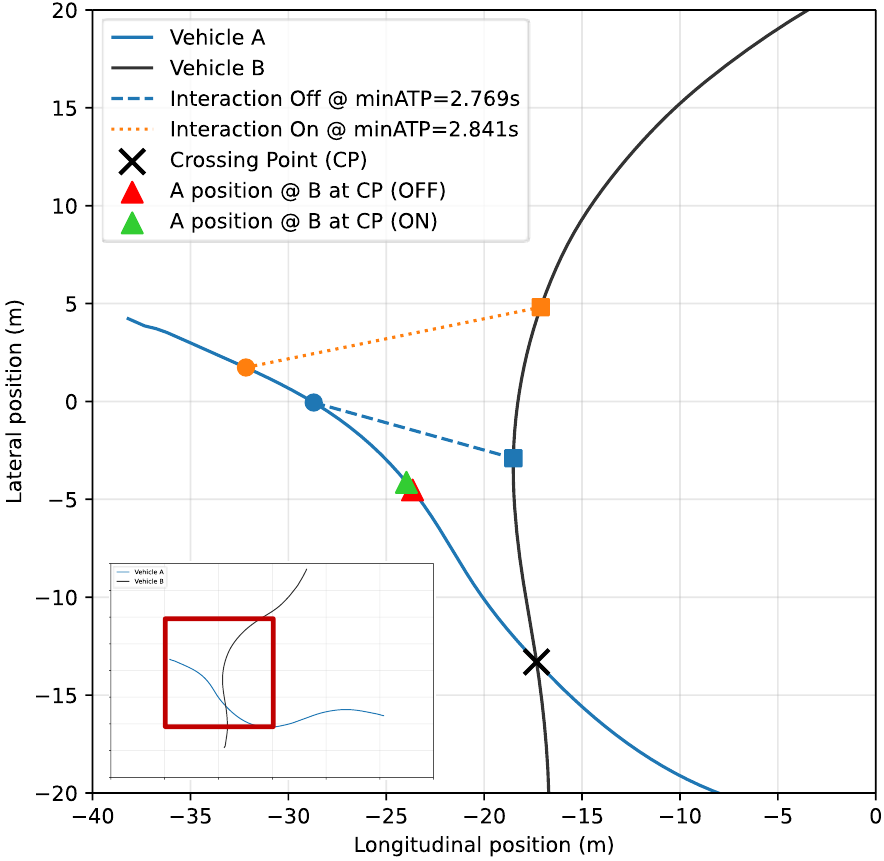}
    \end{subfigure}
    
    \caption{Visualization of Representative Generated Scenarios Comparing Interaction-Off and Interaction-On.}
    \label{fig:interaware_grid}
\end{figure}

Fig.~\ref{fig:interaware_grid} visualizes four representative generated scenarios. While the route geometries remain identical, the spatial arrangements of vehicle~A near the crossing point differ between Interaction-Off and Interaction-On variants. The blue and orange dashed lines indicate the relative positioning of both vehicles at the moment of the minATP for each variant. Furthermore, the red and green triangle markers denote the positions of Vehicle~A at the moment Vehicle~B reaches the crossing point. In the safety-critical scenario, the Interaction-On variant delays Vehicle~A, resulting in a larger minATP and visibly increased clearance at the critical moment, consistent with yielding behavior in roundabouts. In less safety-critical scenarios where yielding is less imperative, the two dashed lines nearly overlap.

Table~\ref{table_intercomparison} summarizes the evaluation of 100 scenarios across two predefined minATP bands. Within the safety-critical band of $[0, 2]$\,s, enabling interaction-aware generation increases the mean minATP from $0.993$\,s to $1.841$\,s and the average clearance from $6.25$\,m to $8.94$\,m. Conversely, in the less critical band of $[2, 4]$\,s, these increments are notably smaller. These results demonstrate that the generator produces interaction-aware behavior that adaptively modulates the safety margin.

\begin{table}[!t]
    \renewcommand{\arraystretch}{1.2}
    \caption{Comparison of Safety Metrics between Interaction-Off and Interaction-On.}
    \label{table_intercomparison}
    \centering
    \begin{tabular*}{0.95\columnwidth}{@{\extracolsep{\fill}} c cc cc @{}}
        \toprule
        \textbf{Predefined} & \multicolumn{2}{c}{\textbf{Mean minATP (s)}} & \multicolumn{2}{c}{\textbf{Mean Clearance (m)}} \\
        \cmidrule(r){2-3} \cmidrule(l){4-5}
        \textbf{minATP Band (s)} & Off & On & Off & On \\
        \midrule
        0\; --\; 2 & 0.993 & 1.841 & 6.253 & 8.938 \\
        2\; --\; 4 & 2.968 & 3.265 & 13.800 & 14.560 \\
        \bottomrule
    \end{tabular*}
\end{table}

\subsection{Criticality-Calibrated Scaling via Interaction Intensity}
\label{sec:results_safety}

While the previous section focuses on interaction-aware scenario generation and qualitative plausibility under a fixed alignment, this section evaluates interaction intensity adjustment as a scalable safety testing intervention. For each trial, ten target minATP values within the safety-critical range ($[0,2]$\,s) are uniformly sampled with step $0.2$\,s. The $\Delta t$ is selected by grid search such that the Interaction-Off scenario reaches a predefined target value minATP rather than a band. A tolerance of $0.05\,\mathrm{s}$ is permitted to improve search efficiency. The interaction intensity is then swept by increasing $\lambda$ from 0 (Interaction-Off) to 1 (full Interaction-On) under the same $\Delta t$.

\begin{figure}[t]
    \centering
    \includegraphics[width=0.4\textwidth]{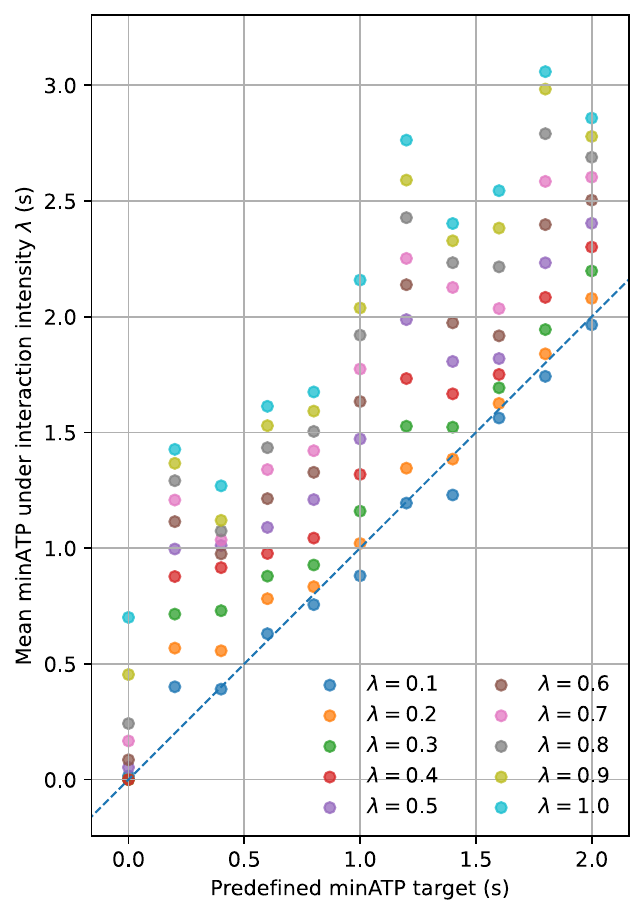}
    \caption{minATP Scatter Plot under Interaction-intensity Sweeps $\lambda$ with Identical Alignment $\Delta t$. The x-axis is the predefined minATP target used to calibrate $\Delta t$ in the Interaction-Off reference, and the y-axis reports the resulting minATP at intensity $\lambda$.}
    \label{fig:safety_main}
\end{figure}

Fig.~\ref{fig:safety_main} presents the minATP scatter plot under identical $\Delta t$. The fact that most points lie above the diagonal indicates that increasing interaction intensity generally increases minATP. A small fraction of points fall below the diagonal due to finite-sample variability and the discrete temporal sampling inherent in the minATP calculation. This capability facilitates effective modulation of criticality without the need for redundant alignment searches.

\section{Conclusion}
This paper proposes an interaction-aware framework for roundabout scenario generation and criticality-calibrated testing. Interaction is represented by a compact yield code based on ATP, with its intensity continuously adjustable via a scalar $\lambda$. A latent-space approach is adopted, where route geometry and temporal progress profiles are mapped to latent codes by pretrained autoencoders and subsequently generated using conditional WGAN-GP models.

Experimental results indicate that the generated roundabout routes adhere to the dominant geometric patterns observed in the dataset. Compared to the baseline model, the interaction-aware TimeWGAN enhances temporal-latent distributional fidelity. The generated scenarios exhibit appropriate yielding behavior, characterized by pronounced responses in safety-critical situations and weak adjustments when yielding is unnecessary. Furthermore, systematically sweeping $\lambda$ tends to increase minATP. This capability facilitates fine-grained, large-scale testing to probe safety boundaries and investigate the optimal execution timing for intelligent driving functions. A limitation is that the current interaction mechanism focuses on yielding during the approach-to-entry segment, while richer multi-agent interaction dynamics remain unexplored. Future work will extend the conditioning beyond yielding to additional interaction modes and investigate critical scenario generation in more complex multi-lane roundabouts.

\bibliographystyle{IEEEtran}
\bibliography{refs}

@article{chu2026imperfect_adas,
    title = {Imperfect advanced driver assistance systems in the eyes of imperfect users},
    journal = {Transportation Research Part F: Traffic Psychology and Behaviour},
    volume = {116},
    pages = {103447},
    year = {2026},
    issn = {1369-8478},
    doi = {https://doi.org/10.1016/j.trf.2025.103447},
    author = {Yueying Chu and Wenting Tang and Shanguang Chen and Peng Liu}
}

@article{kalra2016driving_to_safety,
title = {Driving to safety: How many miles of driving would it take to demonstrate autonomous vehicle reliability?},
journal = {Transportation Research Part A: Policy and Practice},
volume = {94},
pages = {182-193},
year = {2016},
issn = {0965-8564},
doi = {https://doi.org/10.1016/j.tra.2016.09.010},
author = {Nidhi Kalra and Susan M. Paddock}
}

@article{chen2024data_driven_traffic_sim_review,
  title={Data-driven traffic simulation: A comprehensive review},
  author={Chen, Di and Zhu, Meixin and Yang, Hao and Wang, Xuesong and Wang, Yinhai},
  journal={IEEE Transactions on Intelligent Vehicles},
  volume={9},
  number={4},
  pages={4730--4748},
  year={2024},
  publisher={IEEE}
}

@article{ding2023survey_scengen,
  title={A survey on safety-critical driving scenario generation—a methodological perspective},
  author={Ding, Wenhao and Xu, Chejian and Arief, Mansur and Lin, Haohong and Li, Bo and Zhao, Ding},
  journal={IEEE Transactions on Intelligent Transportation Systems},
  volume={24},
  number={7},
  pages={6971--6988},
  year={2023},
  publisher={IEEE}
}

@article{Polders2013Crash,
  title={Identifying Crash Patterns on Roundabouts: An Exploratory Study},
  author={Polders, Evelien and Daniels, Stijn and Casters, Winfried and Brijs, Tom},
  journal={Traffic injury prevention},
  volume={16},
  number={2},
  pages={202--207},
  year={2015},
  publisher={Taylor \& Francis}
}

@article{li2023roundabout_conflict_review,
title = {The role of traffic conflicts in roundabout safety evaluation: A review},
journal = {Accident Analysis \& Prevention},
volume = {196},
pages = {107430},
year = {2024},
issn = {0001-4575},
doi = {https://doi.org/10.1016/j.aap.2023.107430},
author = {Li Li and Zai Zhang and Zhi-Gang Xu and Wen-Chen Yang and Qing-Chang Lu}
}

@InProceedings{arjovsky2017wgan,
  title = 	 {{W}asserstein Generative Adversarial Networks},
  author =       {Martin Arjovsky and Soumith Chintala and L{\'e}on Bottou},
  booktitle = 	 {Proceedings of the 34th International Conference on Machine Learning},
  pages = 	 {214--223},
  year = 	 {2017},
  editor = 	 {Precup, Doina and Teh, Yee Whye},
  volume = 	 {70},
  series = 	 {Proceedings of Machine Learning Research},
  month = 	 {06--11 Aug},
  publisher =    {PMLR}
}

@article{gulrajani2017wgangp,
  title={Improved training of wasserstein gans},
  author={Gulrajani, Ishaan and Ahmed, Faruk and Arjovsky, Martin and Dumoulin, Vincent and Courville, Aaron C},
  journal={Advances in neural information processing systems},
  volume={30},
  year={2017}
}

@techreport{nhtsa2024_ncap_adas_roadmap,
  author       = {{National Highway Traffic Safety Administration}},
  title        = {{New Car Assessment Program Final Decision Notice: Advanced Driver Assistance Systems Roadmap}},
  institution  = {U.S. Department of Transportation},
  year         = {2024},
  month        = nov,
  note         = {Final Decision Notice (NCAP ADAS Roadmap), published 18 Nov 2024}
}

@inproceedings{fu2020_aeb_sdt,
    author = {Fu, Ernestine and Johns, Mishel and Hyde, David A. B. and Sibi, Srinath and Fischer, Martin and Sirkin, David},
    title = {Is Too Much System Caution Counterproductive? Effects of Varying Sensitivity and Automation Levels in Vehicle Collision Avoidance Systems},
    year = {2020},
    isbn = {9781450367080},
    publisher = {Association for Computing Machinery},
    address = {New York, NY, USA},
    doi = {10.1145/3313831.3376300},
    booktitle = {Proceedings of the 2020 CHI Conference on Human Factors in Computing Systems},
    pages = {1–13},
    numpages = {13},
    location = {Honolulu, HI, USA},
    series = {CHI '20}
}

@article{feng2022tslg_adaptive,
  author={Feng, Shuo and Feng, Yiheng and Sun, Haowei and Zhang, Yi and Liu, Henry X.},
  journal={IEEE Transactions on Intelligent Transportation Systems}, 
  title={Testing Scenario Library Generation for Connected and Automated Vehicles: An Adaptive Framework}, 
  year={2022},
  volume={23},
  number={2},
  pages={1213-1222},
  doi={10.1109/TITS.2020.3023668}}

@article{feng2021idit,
  author  = {Feng, Shuo and Yan, Xintao and Sun, Haowei and Feng, Yiheng and Liu, Henry X.},
  title   = {Intelligent driving intelligence test for autonomous vehicles with naturalistic and adversarial environment},
  journal = {Nature Communications},
  year    = {2021},
  volume  = {12},
  number  = {1},
  pages   = {1--14},
  doi     = {10.1038/s41467-021-21007-8}
}

@article{liu2024curse,
  author  = {Liu, Henry X. and Feng, Shuo},
  title   = {Curse of rarity for autonomous vehicles},
  journal = {Nature Communications},
  year    = {2024},
  volume  = {15},
  number  = {1},
  pages   = {4808},
  doi     = {10.1038/s41467-024-49194-0}
}

@article{degelder2022sr,
  author={de Gelder, Erwin and Hof, Jasper and Cator, Eric and Paardekooper, Jan-Pieter and Camp, Olaf Op den and Ploeg, Jeroen and de Schutter, Bart},
  journal={IEEE Transactions on Intelligent Transportation Systems}, 
  title={Scenario Parameter Generation Method and Scenario Representativeness Metric for Scenario-Based Assessment of Automated Vehicles}, 
  year={2022},
  volume={23},
  number={10},
  pages={18794-18807},
  doi={10.1109/TITS.2022.3154774}
}

@article{li2025multi,
  title={Multi-Agent Scenario Generation in Roundabouts with a Transformer-enhanced Conditional Variational Autoencoder},
  author={Li, Li and Brinkmann, Tobias and Temmen, Till and Eisenbarth, Markus and Andert, Jakob},
  journal={arXiv preprint arXiv:2510.24671},
  year={2025}
}

@article{li2023aap_wgangp,
title = {Data generation for connected and automated vehicle tests using deep learning models},
journal = {Accident Analysis \& Prevention},
volume = {190},
pages = {107192},
year = {2023},
issn = {0001-4575},
doi = {https://doi.org/10.1016/j.aap.2023.107192},
author = {Ye Li and Fei Liu and Lu Xing and Yi He and Changyin Dong and Chen Yuan and Jiguang Chen and Lu Tong}
}

@article{liu2025eaai_ttsgan,
title = {Generating intersection pre-crash trajectories for autonomous driving safety testing using Transformer Time-Series Generative Adversarial Networks},
journal = {Engineering Applications of Artificial Intelligence},
volume = {160},
pages = {111995},
year = {2025},
issn = {0952-1976},
doi = {https://doi.org/10.1016/j.engappai.2025.111995},
author = {Xichang Liu and Helai Huang and Jiang Bian and Rui Zhou and Zhiyuan Wei and Hanchu Zhou}
}

@inproceedings{xu2025diffscene,
author = {Xu, Chejian and Petiushko, Aleksandr and Zhao, Ding and Li, Bo},
title = {DiffScene: diffusion-based safety-critical scenario generation for autonomous vehicles},
year = {2025},
isbn = {978-1-57735-897-8},
publisher = {AAAI Press},
doi = {10.1609/aaai.v39i8.32951},
booktitle = {Proceedings of the Thirty-Ninth AAAI Conference on Artificial Intelligence and Thirty-Seventh Conference on Innovative Applications of Artificial Intelligence and Fifteenth Symposium on Educational Advances in Artificial Intelligence},
articleno = {978},
numpages = {9},
series = {AAAI'25/IAAI'25/EAAI'25}
}

@article{ma2024evolving_trc,
title = {Evolving testing scenario generation and intelligence evaluation for automated vehicles},
journal = {Transportation Research Part C: Emerging Technologies},
volume = {163},
pages = {104620},
year = {2024},
issn = {0968-090X},
doi = {https://doi.org/10.1016/j.trc.2024.104620},
author = {Yining Ma and Wei Jiang and Lingtong Zhang and Junyi Chen and Hong Wang and Chen Lv and Xuesong Wang and Lu Xiong}
}

@article{bondtaylor2022dgm,
  author={Bond-Taylor, Sam and Leach, Adam and Long, Yang and Willcocks, Chris G.},
  journal={IEEE Transactions on Pattern Analysis and Machine Intelligence}, 
  title={Deep Generative Modelling: A Comparative Review of VAEs, GANs, Normalizing Flows, Energy-Based and Autoregressive Models}, 
  year={2022},
  volume={44},
  number={11},
  pages={7327-7347},
  doi={10.1109/TPAMI.2021.3116668}
}

@article{cao2024diffusion,
  author={Cao, Hanqun and Tan, Cheng and Gao, Zhangyang and Xu, Yilun and Chen, Guangyong and Heng, Pheng-Ann and Li, Stan Z.},
  journal={IEEE Transactions on Knowledge and Data Engineering}, 
  title={A Survey on Generative Diffusion Models}, 
  year={2024},
  volume={36},
  number={7},
  pages={2814-2830},
  doi={10.1109/TKDE.2024.3361474}
}

@article{hayward1972near_miss,
  author      = {Hayward, John C.},
  title       = {Near Miss Determination Through Use of a Scale of Danger},
  journal     ={Highway Research Record},
  year        = {1972},
}

@article{peesapati2018pet,
title = {Can post encroachment time substitute intersection characteristics in crash prediction models?},
journal = {Journal of Safety Research},
volume = {66},
pages = {205-211},
year = {2018},
issn = {0022-4375},
doi = {https://doi.org/10.1016/j.jsr.2018.05.002},
author = {Lakshmi N. Peesapati and Michael P. Hunter and Michael O. Rodgers}
}

@article{astarita2019surrogate,
  title={Surrogate safety measures from traffic simulation models a comparison of different models for intersection safety evaluation},
  author={Astarita, Vittorio and Festa, Demetrio Carmine and Giofr{\`e}, Vincenzo Pasquale and Guido, Giuseppe},
  journal={Transportation research procedia},
  volume={37},
  pages={219--226},
  year={2019},
  publisher={Elsevier}
}

@article{memeh2024roundabout_gap_crash,
title = {Gap acceptance behaviour and crash risks of mobile phone distracted young drivers at roundabouts: A random parameters survival model},
journal = {Accident Analysis \& Prevention},
volume = {206},
pages = {107720},
year = {2024},
issn = {0001-4575},
doi = {https://doi.org/10.1016/j.aap.2024.107720},
author = {Esther Memeh and Yasir Ali and Francisco {Javier Rubio} and Craig Hancock and Md {Mazharul Haque}}
}

@article{zhao2018carfollow,
  title={Accelerated evaluation of automated vehicles in car-following maneuvers},
  author={Zhao, Ding and Huang, Xianan and Peng, Huei and Lam, Henry and LeBlanc, David J},
  journal={IEEE Transactions on Intelligent Transportation Systems},
  volume={19},
  number={3},
  pages={733--744},
  year={2017},
  publisher={IEEE}
}

@article{sun2022doe,
  author={Sun, Jian and Zhou, Huajun and Xi, Haochen and Zhang, He and Tian, Ye},
  journal={IEEE Transactions on Intelligent Transportation Systems}, 
  title={Adaptive Design of Experiments for Safety Evaluation of Automated Vehicles}, 
  year={2022},
  volume={23},
  number={9},
  pages={14497-14508},
  doi={10.1109/TITS.2021.3130040}
}

@article{zhu2022hazard,
  author={Zhu, Bing and Zhang, Peixing and Zhao, Jian and Deng, Weiwen},
  journal={IEEE Transactions on Intelligent Transportation Systems}, 
  title={Hazardous Scenario Enhanced Generation for Automated Vehicle Testing Based on Optimization Searching Method}, 
  year={2022},
  volume={23},
  number={7},
  pages={7321-7331},
  doi={10.1109/TITS.2021.3068784}
}

@INPROCEEDINGS{krajewski2020round,
  author={Krajewski, Robert and Moers, Tobias and Bock, Julian and Vater, Lennart and Eckstein, Lutz},
  booktitle={2020 IEEE 23rd International Conference on Intelligent Transportation Systems (ITSC)}, 
  title={The rounD Dataset: A Drone Dataset of Road User Trajectories at Roundabouts in Germany}, 
  year={2020},
  volume={},
  number={},
  pages={1-6},
  doi={10.1109/ITSC45102.2020.9294728}
}

\end{document}